\documentclass[sigconf]{acmart}

\usepackage{tabularx}
\usepackage{adjustbox}
\usepackage{multicol}
\usepackage{rotating}
\usepackage{hyperref}
\usepackage{multirow}
\usepackage{graphicx}
\usepackage{booktabs}
\usepackage{graphics}
\usepackage{xspace} 
\usepackage{color,soul}
\usepackage{booktabs}
\usepackage{mdwlist}
\usepackage{multirow}
\usepackage{amsmath}
\usepackage{enumitem} 
\usepackage{calc}
\usepackage{colortbl}
\usepackage[ruled,linesnumbered]{algorithm2e}
\usepackage{booktabs}
\usepackage{mathtools}
\usepackage{balance}
\usepackage{comment}
\usepackage{textcomp}
\usepackage{subcaption}
\usepackage{tcolorbox}
\usepackage{listings}
\usepackage{xcolor}
\usepackage{colortbl}

\AtBeginDocument{%
  }

\setcopyright{acmlicensed}
\copyrightyear{2018}
\acmYear{2018}
\acmDOI{XXXXXXX.XXXXXXX}

\acmConference[Conference acronym 'XX]{Make sure to enter the correct
  conference title from your rights confirmation emai}{June 03--05,
  2018}{Woodstock, NY}
\acmISBN{978-1-4503-XXXX-X/18/06}

\usepackage{tcolorbox}
\newtcolorbox{mypromptbox}[1][]{colback=lightgray!10, colframe=gray!75, breakable, title=Prompt}
\tcbuselibrary{breakable} 

\usepackage{booktabs}
\usepackage{multirow} 
\usepackage{array} 

\copyrightyear{2024}
\acmYear{2024}
\setcopyright{rightsretained}
\acmConference[JCDL '24]{The 2024 ACM/IEEE Joint Conference on Digital Libraries}{December 16--20, 2024}{Hong Kong, China}
\acmBooktitle{The 2024 ACM/IEEE Joint Conference on Digital Libraries (JCDL '24), December 16--20, 2024, Hong Kong, China}\acmDOI{10.1145/3677389.3702601}
\acmISBN{979-8-4007-1093-3/24/12}

\begin{document}

\title[In Quest for Political Truth]{Navigating Nuance: In Quest for Political Truth}

\author{Soumyadeep Sar}
\affiliation{%
  \institution{Indian Institute of Science Education and Research}
  \city{Kolkata}
  \country{India}
}
\email{soumyadeepsar26@gmail.com}

\author{Dwaipayan Roy}
\affiliation{%
  \institution{Indian Institute of Science Education and Research}
  \city{Kolkata}
  \country{India}
}
\email{dwaipayan.roy@iiserkol.ac.in}

%
\renewcommand{\shortauthors}{Soumyadeep Sar, \& Dwaipayan Roy}








\begin{abstract}
This study investigates the several nuanced rationales for countering the rise of political bias. 
We evaluate the performance of the Llama-3 (70B) language model on the Media Bias Identification Benchmark (MBIB), based on a novel prompting technique that incorporates subtle reasons for identifying political leaning. 
Our findings underscore the challenges of detecting political bias and highlight the potential of transfer learning methods to enhance future models. 
Through our framework, we achieve a comparable performance with the supervised and fully fine-tuned ConvBERT model, which is the state-of-the-art model, performing best among other baseline models for the political bias task on MBIB.
By demonstrating the effectiveness of our approach, we contribute to the development of more robust tools for mitigating the spread of misinformation and polarization. Our codes and dataset are made publicly available in github\footnote{\url{https://github.com/Soumyadeepsar/Navigating-Nuance-In-Quest-for-Political-Truth}}.

\end{abstract}

\ccsdesc[300]{Information systems~Social recommendation}

\keywords{LLMs, Prompting, Bias Detection, Chain-of-Thought.}

\maketitle

\section{Introduction}

Political bias detection has become a critical area of research in natural language processing (NLP) due to its significant influence on media literacy, public opinion, and democratic processes. The pervasive nature of political bias in media and online content demands robust methodologies for its identification and analysis. Political bias can manifest in various forms, including word choice, framing of issues, and selective omission of information, all of which can subtly influence readers' perceptions and beliefs (Entman, 2007)~\cite{article}.
Recent advancements in machine learning and NLP have enabled the development of sophisticated models for detecting political bias. Traditional approaches relied heavily on lexicon-based methods, which involved predefined lists of biased terms and phrases. While these methods provided a foundation, they often struggled with the nuanced and context-dependent nature of political bias. More contemporary techniques leveraging deep learning and large language models (LLMs) to capture subtler forms of bias and context-specific variations in sentence-level text have been done~\cite{DBLP:conf/emnlp/HongCHJT23}. Efforts have been made to learn the factuality of reporting and bias, trying to categorize entire news media based upon different features collected through its URLS, websites, Wikipedia page, Twitter account and many other factors~\cite{DBLP:conf/emnlp/BalyKAGN18}.

One promising approach in the domain of in-context learning is the use of Chain-of-Thought (CoT) prompting, which involves generating and utilizing intermediate reasoning steps to improve the reasoning performance of LLMs. This method aims to enhance the model's interpretability and accuracy by breaking down the decision-making process into a sequence of logical steps~\cite{DBLP:conf/nips/Wei0SBIXCLZ22}. By employing CoT prompting, researchers can better understand the underlying reasoning of LLMs and ensure more reliable and better performance on complex tasks. Hence we try to leverage COT, to shape proper reasoning and understanding of political bias in LLMs, resulting in better performance. 

Many studies were conducted in past for learning connections and patterns in biased text, trying to devise automatic detection solutions~\cite{DBLP:conf/coling/Perez-RosasKLM18}. Despite these advancements, challenges remain in achieving high accuracy and generalizability across diverse datasets and political contexts. The inherently subjective nature of political bias along with other factors and the dynamic evolution of political discourse, add complexity to this task. Therefore, research is essential to try to put this hypothesis to test and develop standardized techniques for further research in this area.

This paper explores the effectiveness of COT prompting in improving the ability of LLMs to classify statements as biased or unbiased. We build on existing literature and propose a novel prompt that utilizes a COT technique based upon subtle reasoning steps to enhance political bias detection. Through extensive experimentation and analysis, we aim to contribute to the growing body of work in this field.

\section{Related work}
Prompt-based fine-tuning has significantly enhanced the performance of Pre-trained Language Models (PLMs) in few-shot text classification by using task-specific prompts. However, since PLMs are not pre-trained with prompt-style expressions, their effectiveness in few-shot learning is limited. To address this, the Unified Prompt Tuning (UPT) framework was introduced by~\citet{DBLP:conf/emnlp/Wang0LTQYSHG22}, which improves few-shot text classification for BERT-style models by capturing prompting semantics from various non-target NLP datasets. UPT employs a novel Prompt-Options-Verbalizer paradigm for joint prompt learning across tasks, enabling PLMs to acquire task-invariant prompting knowledge. Additionally, a self-supervised task, Knowledge-enhanced Selective Masked Language Modeling, is designed to boost the PLM’s generalization capabilities, allowing better adaptation to unfamiliar tasks in low-resource settings. Experimental results across diverse NLP tasks demonstrate that UPT consistently outperforms existing state-of-the-art methods in prompt-based fine-tuning.

The study by~\citet{Wen_2023} investigates the ability of large language models (LLMs), such as ChatGPT, to detect media bias by utilizing the Media Bias Identification Benchmark (MBIB)~\cite{Wessel2023}. ChatGPT's performance is compared against fine-tuned models like Bidirectional and AutoRegressive Transformers (BART)~\cite{lewis-etal-2020-bart}, Convolutional Bidirectional Encoder Representations from Transformers (ConvBERT)~\cite{jiang2021convbertimprovingbertspanbased}, and Generative Pre-trained Transformer 2 (GPT-2). The results reveal that while ChatGPT is on par with fine-tuned models in identifying hate speech and text-level context bias, it struggles with detecting subtler elements of media bias, including fake news, racial, gender, political and cognitive biases.

Large language models (LLMs) have excelled in tasks such as dialogue generation, commonsense reasoning, and question-answering. In-context learning (ICL) is a key method for adapting LLMs to downstream tasks through prompt-based demonstrations. However, the performance can be significantly impacted by the distribution of these demonstrations, particularly in challenging classification tasks. \citet{du-etal-2023-task} introduce task-level thinking steps to mitigate bias introduced by demonstrations. Additionally, they propose a progressive revision framework to enhance thinking steps by correcting difficult demonstrations. Experimental results demonstrate that their method outperforms others on three types of challenging classification tasks in both zero-shot and few-shot settings. The incorporation of task-level thinking steps and automatically generated CoT yields more competitive performance. 

\begin{figure}[t]
    \centering
    \includegraphics[width=0.38\textwidth]{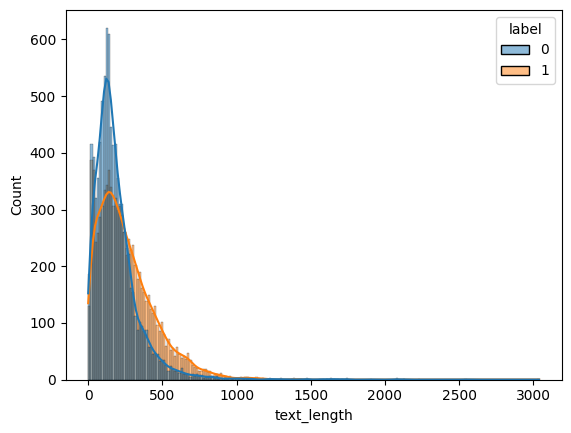}
    \caption{Distribution of text according to length}
    \label{fig:distribution}
\end{figure}

\section{Dataset}
 Media bias detection is a complex multi-task problem which can be very challenging on its own. There is a wide variety of bias propagating in various news outlet and media platforms. The Media Bias Identification Benchmark (MBIB)~\cite{Wessel2023}, provides an extensive benchmark that groups different types of media bias (e.g., linguistic, cognitive, political) under a common framework which allows researchers to develop and test their techniques against different types of bias detection. Notably, this is the first-ever large benchmark-level task designed specifically for bias detection. 

The Dataset is well-balanced, with an equal number of biased and unbiased text. The frequency distribution of statements according to text length across both labels is illustrated in Figure~\ref{fig:distribution}.


\section{Methodology }
\subsection{Data-preprocessing}
For the experiments, we use the dataset from mbib-base\footnote{\url{https://huggingface.co/datasets/mediabiasgroup/mbib-base}}. 
It has 9 different splits consisting of all the different types of bias the benchmark was designed for. 
Among these, the political bias dataset consists of 17,704 data points.
The dataset begins with all the unbiased texts 8852 (50\% of the dataset), grouped together, at the end of all unbiased text we had the entire biased statements. 
To mix up the data points, we randomly shuffle the dataset using the seed value 42 in Huggingface's \textit{datasets} function, \texttt{shuffle()}\footnote{\url{https://huggingface.co/docs/datasets/en/process\#shuffle}}, to ensure reproducibility of the intermixed data. 
We then divide the shuffled dataset into 18 equal-sized chunks, where each chunk has around 1000 statements.
Due to the random shuffling of the dataset, every chunk remains well-balanced in terms of the two class distributions.

\begin{figure}[t]
    \centering
    \includegraphics[width=0.42\textwidth]{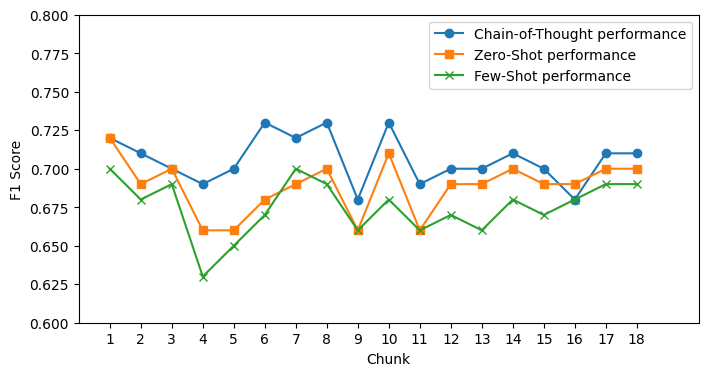} 
    \caption{Macro-F1 score across 18 chunks for different prompt techniques.}
    \label{fig:performance}
\end{figure}

\begin{table*}[t]
  \caption{Misclassified Examples used in Chain-of-Thought Prompt}
  \label{tab:examples}
  \centering
  \resizebox{0.8\textwidth}{!}{
  \begin{tabular}{p{3cm}p{12cm}p{2cm}} 
    \toprule
    Prompting Techniques & Examples & Actual label \\
    \midrule
    \multirow{6}{3cm}{Chain-of-Thought} & \texttt{In the video clip above, Bernice apologizes about the entire social media fiasco between her and Tiny, stating "I’m very disappointed in what I did.}"& unbiased (0) \\ 
    & \\
    & \texttt{Statement: Fox News James Rosen and Jake Gibson recently reported the wife of Justice Department official Bruce G Ohr worked for the opposition research firm during the 2016 presidential election.}& biased (1) \\
    \bottomrule
  \end{tabular}
  }
\end{table*}

\subsection{Large language model}
We use the Open source model Llama-3-70b~\cite{llama3modelcard} by Meta, for all the prompting experiments. 
Llama-3 performs better than its predecessors and rival LLMs across various benchmarks, such as MMLU~\cite{DBLP:journals/corr/abs-2009-03300} and HumanEval~\cite{chen2021codex}. 
Llama 3 has outperformed other high-parameter models like Google’s Gemini 1.5 Pro~\cite{DBLP:journals/corr/abs-2403-05530} and Anthropic’s Claude 3 Sonnet~\cite{Anthropic}, especially in complex reasoning and comprehension tasks; this particularly makes Llama-3 a good choice for high-order reasoning tasks.
We perform zero-shot, and few-shot prompting, together with an advanced variant of Chain-of-Thought prompting. 
To make our model lightweight, we used an API service from the Groq platform for our experiments. This reduces the space overhead.
Particularly, we used the Langchain-groq\footnote{\url{https://github.com/langchain-ai/langchain/tree/master/libs/partners/groq}} integration to prompt Llama-3-70b-Instruct model\footnote{\url{https://huggingface.co/meta-llama/Meta-Llama-3-70B-Instruct}}. The \textit{temperature} of the model was set to \textit{0.0}, to prevent it from hallucinating and deviating from instructions passed in the prompts. All the other parameters were set to default settings.

\subsection{Selection of examples and prompt design}

A straightforward method we employ is zero-shot prompting, which involves directly providing the model with a task or question without the need for any task-specific training or examples beforehand. 
The specific prompt used for this approach is presented in Appendic~\ref{appendix:zero}.
The next approach, few-shot~\cite{brown2020languagemodelsfewshotlearners} prompt, was constructed with 8 examples randomly selected from the entire dataset. 
An equal number of biased and unbiased statements were chosen to ensure fair representation of both classes. While selecting samples, we designated a unique seed value (42) to ensure the reproducibility of the prompt. The template can be seen in Appendix~\ref{appendix:few}.
The selected examples contained single-word text as well, which gives the model an example of how to label short texts that are contextually inconclusive and might confuse the model. 

To instigate the model with nuanced ways of thinking, we implement a k-shot Chain-of-Thought prompt setting.
We selected two examples ($k=2$) based on a specific rationale for the CoT prompt (detailed in Appendix~\ref{appendix:cot}). These examples were chosen from Chunk 8 after running the zero-shot prompt. We carefully observed the misclassified statements, which clearly indicated where the model struggled to reason correctly, leading to incorrect labels. Consequently, we selected two such examples, as shown in Table~\ref{tab:examples}. We then provide nuanced and subtle reasoning steps in the prompt involving understanding agenda, omission of facts, and other steps to instigate better thinking in the LLM. The complete prompt is given in Appendix~\ref{appendix:cot}, where all the steps are used to guide the model towards better performance can be seen.

    







\section{Result}
We provide the entire performance distribution by LLama-3 based on different types of prompt settings in Table~\ref{tab:results}. The performance of each prompting technique across all 18 chunks is graphically illustrated in Figure~\ref{fig:performance}. We can see a good amount of performance boost in Chunks 4, 5, 6, 7, 8, 9, and 11 via our designed k-shot Chain-of-Thought prompt. We also see a nominal increase in the macro-F1 score in the other remaining chunks. The few-shot settings underperformed in most of the chunks compared to Chain-of-Thought and even zero-shot prompts. This can be attributed to the fact that the few-shot heavily relies on the selection of examples for the in-context learning of the model, which may sometimes mislead the LLM, or may not instil deeper reasoning behind the task. 
Overall, we noticed that the CoT prompting achieved the best performance across the chunks. 

However, few-shot prompting had the advantage of training the model to follow user instructions strictly, returning only the label (0 or 1) without additional information. In contrast, the CoT approach occasionally deviated by providing brief explanations along with the label, despite explicit instructions to avoid doing so.
The results show that the performance differences between the various prompt-based methods across the chunks are relatively consistent.
Interestingly, the zero-shot method performs better than the few-shot counterpart.
This is due to the bias inherent in the data as well as the example using which the few-shot model is trained. 
Further, the Chain-of-Thought achieves the best performance although the difference in performance is minimal.
Note that, the best-performing baseline, i.e. ConvBERT (as reported in~\cite{Wessel2023}) achieves an average Macro-F1 score of $0.7110$.
We are achieving performance on par with the prompt-based CoT settings when compared to the ConvBERT model, which is a fully supervised model. 
This is particularly noteworthy given that the CoT model operates in a prompting framework, relying only on in-context learning rather than the extensive labelled data and fine-tuning that the ConvBERT model requires.
The fact that a prompting-based method can match the performance of a supervised model shows the potential and effectiveness of CoT prompting, especially in tasks where reasoning and context play a significant role.
\section{Conclusion and future directions}
With the advent of new larger language models, it is increasingly becoming resource-intensive to train such extremely large models on our task-oriented dataset. So in-context learning techniques serve as an excellent means to teach the model how to perform through proper contextual examples and instructions. In this paper, we observed the performance of one of the best open-source models (Llama-3) on bias detection, which is a very alarming problem in this era of social media. We see it performs at par with the supervised and fine-tuned baseline (ConvBERT). 
 
The study reveals the complexity of political bias creeping into our social media. We also saw subjectivity does play a vital role in the propagation of political bias, but there are many other factors, like news report outlets, fact selection, and omission of context, that was considered while designing the final improvised version of the CoT prompt. Hence, significant research is needed in this field, to acquire better and accurate insights to solve the problem of political bias.

\begin{table}[t]
  \caption{
  The Macro-F1 is reported for each chunk by respective methods. The best performance for each chunk in bold.}
  \label{tab:results}
  \centering
  \begin{adjustbox}{width=0.9\columnwidth}
  \begin{tabular}{cccc} 
    \toprule
    Chunk no. & Chain-of-Thought & Zero-shot & Few-shot \\ 
    \midrule
    1 & \textbf{0.72} & 0.72 & 0.70 \\
    2 & \textbf{0.71} & 0.69 & 0.68 \\
    3 & \textbf{0.70} & 0.70 & 0.69 \\
    4 & \textbf{0.69} & 0.66 & 0.63 \\
    5 & \textbf{0.70} & 0.66 & 0.65 \\
    6 & \textbf{0.73} & 0.68 & 0.67 \\
    7 & \textbf{0.72} & 0.69 & 0.70 \\
    8 & \textbf{0.73} & 0.70 & 0.69 \\
    9 & \textbf{0.68} & 0.66 & 0.66 \\
    10 & \textbf{0.73} & 0.71 & 0.68 \\
    11 & \textbf{0.69} & 0.66 & 0.66 \\
    12 & \textbf{0.70} & 0.69 & 0.67 \\
    13 & \textbf{0.70} & 0.69 & 0.66 \\
    14 & \textbf{0.71} & 0.70 & 0.68 \\
    15 & \textbf{0.70} & 0.69 & 0.67 \\
    16 & 0.68 & \textbf{0.69} & 0.68 \\ 
    17 & \textbf{0.71 }& 0.70 & 0.69 \\ 
    18 & \textbf{0.71} & 0.70 & 0.69 \\ 
    \midrule
    Average (entire dataset) & \textbf{0.7061} & 0.6883 & 0.6749 \\
    \bottomrule
  \end{tabular}
  \end{adjustbox}
\end{table}




\begin{appendix}
\section{Appendix}

In this section, we outline the prompts used at different stages of our pipeline. 
Specifically, we detail how we employed \textit{zero-shot}, \textit{few-shot}, and \textit{Chain-of-Thought} prompting strategies. These approaches are presented sequentially, as each serves a distinct purpose within our process.
\subsection{Zero-Shot Prompt}\label{appendix:zero}
\begin{tcolorbox}[colback=lightgray!10, colframe=gray!75, title=Prompt used for zero-shot]
Classify the following statement as politically biased or unbiased. Give the final label only as 0 (unbiased) or 1 (biased), do not write your explanation.

\textbf{Statement: }\{input\}

\textbf{Label:}
\end{tcolorbox}

\subsection{Few-Shot Prompt}\label{appendix:few}
\begin{tcolorbox}[colback=lightgray!10, colframe=gray!75, title=Prompt used for few-shot]    
Given an input sentence try to classify whether it is politically biased (1) or unbiased (0). Only give your label as 0 or 1 for the statement. Some examples are given below:

\textbf{Statement:} \{ text\}

\textbf{Label:} 0

......4 pairs of unbiased data points....

\textbf{Statement:} \{ text\}

\textbf{Label:} 1

.........4 pairs of biased data points..........

Give the label for \textbf{Statement:} \{input\}

\textbf{Label:} 
\end{tcolorbox}

\subsection{Chain-of-Thought prompt}\label{appendix:cot}

\begin{tcolorbox}[colback=lightgray!10, colframe=gray!75, title=Prompt used for Chain-of-Thought]
Here are some examples of how to classify a statement as biased or unbiased :

\textbf{Example-1:}
\textbf{Statement:} Refer to Table~\ref{tab:examples} for text


\textbf{Step-by-Step thinking:}

\textbf{1.Fact-based reporting(\textit{Objective nature):}} The text reports on a specific event (the video clip), hence objective in nature.

\textbf{2.Neutral language:} The statement is neutral by nature. The statement also lacks an emotive or sensational tone.

\textbf{3. No implicit or explicit bias:} There's no apparent bias towards or against Bernice, Tiny, or their social media issues. Statement appears to be a factual report of public apology without any apparent political bias or agenda.


\textbf{Label: 0 (unbiased)}

\textbf{Example-2: Statement:} Refer to Table~\ref{tab:examples} for text

\textbf{Step-by-Step thinking:}

\textbf{1.Selection of facts:} The statement selects a specific fact about Bruce Ohr's wife working for an opposition research firm, which might be perceived as cherry-picking information to support a particular narrative.

\textbf{2.Contextual omission:} The statement lacks context about the firm's work, its significance, or Bruce Ohr's role, which might create a misleading impression.

\textbf{3.Implication by association:} The statement shows a connection between Bruce Ohr's wife, the opposition research firm, and biased activities, which could be seen as implication by association.


\textbf{Label: 1 (biased)}

For biased (1) short phrases/text (contextually inconclusive), the following reasons must be checked:

\textbf{1.Use of Emotive language}: Strong emotive, sensational, foul and sarcastic words and phrases can be connected to biased text.

\textbf{2.Opinion-based words(\textit{Subjective}):} Phrases like "Analysis Opinion" contain words. Presence of neutral words.

\textbf{3.Partisan sources:} Phrases mentioning specific news sources like "Fox News reports" may be seen as biased due to the perceived political leanings of those sources.


For unbiased (0) incomplete texts following reasons are crucial:

\textbf{1.\textit{Uses objective tone},focuses on information:} Phrases focus on conveying information. They also avoid loaded terms.

\textbf{2.Report facts:} Many phrases report factual information, such as news headlines, dates, names, and events, neutral in nature, without emotive or sensational tone.

\textbf{3. Cite credible sources.}


Classify the following statement as politically biased or unbiased. Only give your label as 0 (unbiased) or 1 (biased).No need to give your steps of thinking.
\textbf{Statement:\{ input \}}\\
\textbf{Label:}
\end{tcolorbox}

\end{appendix}

\bibliographystyle{ACM-Reference-Format}
\balance
\bibliography{sample-base}


\begin{thebibliography}{17}


\ifx \showCODEN    \undefined \def \showCODEN     #1{\unskip}     \fi
\ifx \showDOI      \undefined \def \showDOI       #1{#1}\fi
\ifx \showISBNx    \undefined \def \showISBNx     #1{\unskip}     \fi
\ifx \showISBNxiii \undefined \def \showISBNxiii  #1{\unskip}     \fi
\ifx \showISSN     \undefined \def \showISSN      #1{\unskip}     \fi
\ifx \showLCCN     \undefined \def \showLCCN      #1{\unskip}     \fi
\ifx \shownote     \undefined \def \shownote      #1{#1}          \fi
\ifx \showarticletitle \undefined \def \showarticletitle #1{#1}   \fi
\ifx \showURL      \undefined \def \showURL       {\relax}        \fi
\providecommand\bibfield[2]{#2}
\providecommand\bibinfo[2]{#2}
\providecommand\natexlab[1]{#1}
\providecommand\showeprint[2][]{arXiv:#2}

\bibitem[AI@Meta(2024)]%
        {llama3modelcard}
\bibfield{author}{\bibinfo{person}{AI@Meta}.} \bibinfo{year}{2024}\natexlab{}.
\newblock \showarticletitle{Llama 3 Model Card}.
\newblock  (\bibinfo{year}{2024}).
\newblock
\urldef\tempurl%
\url{https://github.com/meta-llama/llama3/blob/main/MODEL_CARD.md}
\showURL{%
\tempurl}


\bibitem[Anthropic(2024)]%
        {Anthropic}
\bibfield{author}{\bibinfo{person}{Anthropic}.}
  \bibinfo{year}{2024}\natexlab{}.
\newblock \showarticletitle{Claude-3 Model}.
\newblock  (\bibinfo{year}{2024}).
\newblock
\urldef\tempurl%
\url{https://www-cdn.anthropic.com/de8ba9b01c9ab7cbabf5c33b80b7bbc618857627/Model_Card_Claude_3.pdf}
\showURL{%
\tempurl}


\bibitem[Baly et~al\mbox{.}(2018)]%
        {DBLP:conf/emnlp/BalyKAGN18}
\bibfield{author}{\bibinfo{person}{Ramy Baly}, \bibinfo{person}{Georgi
  Karadzhov}, \bibinfo{person}{Dimitar Alexandrov}, \bibinfo{person}{James~R.
  Glass}, {and} \bibinfo{person}{Preslav Nakov}.}
  \bibinfo{year}{2018}\natexlab{}.
\newblock \showarticletitle{Predicting Factuality of Reporting and Bias of News
  Media Sources}. In \bibinfo{booktitle}{\emph{Proceedings of the 2018
  Conference on Empirical Methods in Natural Language Processing, Brussels,
  Belgium, October 31 - November 4, 2018}},
  \bibfield{editor}{\bibinfo{person}{Ellen Riloff}, \bibinfo{person}{David
  Chiang}, \bibinfo{person}{Julia Hockenmaier}, {and} \bibinfo{person}{Jun'ichi
  Tsujii}} (Eds.). \bibinfo{publisher}{Association for Computational
  Linguistics}, \bibinfo{pages}{3528--3539}.
\newblock
\urldef\tempurl%
\url{https://doi.org/10.18653/V1/D18-1389}
\showDOI{\tempurl}


\bibitem[Brown et~al\mbox{.}(2020)]%
        {brown2020languagemodelsfewshotlearners}
\bibfield{author}{\bibinfo{person}{Tom~B. Brown}, \bibinfo{person}{Benjamin
  Mann}, \bibinfo{person}{Nick Ryder}, \bibinfo{person}{Melanie Subbiah},
  \bibinfo{person}{Jared Kaplan}, \bibinfo{person}{Prafulla Dhariwal},
  \bibinfo{person}{Arvind Neelakantan}, \bibinfo{person}{Pranav Shyam},
  \bibinfo{person}{Girish Sastry}, \bibinfo{person}{Amanda Askell},
  \bibinfo{person}{Sandhini Agarwal}, \bibinfo{person}{Ariel Herbert-Voss},
  \bibinfo{person}{Gretchen Krueger}, \bibinfo{person}{Tom Henighan},
  \bibinfo{person}{Rewon Child}, \bibinfo{person}{Aditya Ramesh},
  \bibinfo{person}{Daniel~M. Ziegler}, \bibinfo{person}{Jeffrey Wu},
  \bibinfo{person}{Clemens Winter}, \bibinfo{person}{Christopher Hesse},
  \bibinfo{person}{Mark Chen}, \bibinfo{person}{Eric Sigler},
  \bibinfo{person}{Mateusz Litwin}, \bibinfo{person}{Scott Gray},
  \bibinfo{person}{Benjamin Chess}, \bibinfo{person}{Jack Clark},
  \bibinfo{person}{Christopher Berner}, \bibinfo{person}{Sam McCandlish},
  \bibinfo{person}{Alec Radford}, \bibinfo{person}{Ilya Sutskever}, {and}
  \bibinfo{person}{Dario Amodei}.} \bibinfo{year}{2020}\natexlab{}.
\newblock \bibinfo{title}{Language Models are Few-Shot Learners}.
\newblock
\newblock
\showeprint[arxiv]{2005.14165}~[cs.CL]
\urldef\tempurl%
\url{https://arxiv.org/abs/2005.14165}
\showURL{%
\tempurl}


\bibitem[Chen et~al\mbox{.}(2021)]%
        {chen2021codex}
\bibfield{author}{\bibinfo{person}{Mark Chen}, \bibinfo{person}{Jerry Tworek},
  \bibinfo{person}{Heewoo Jun}, \bibinfo{person}{Qiming Yuan},
  \bibinfo{person}{Henrique~Ponde de Oliveira~Pinto}, \bibinfo{person}{Jared
  Kaplan}, \bibinfo{person}{Harri Edwards}, \bibinfo{person}{Yuri Burda},
  \bibinfo{person}{Nicholas Joseph}, \bibinfo{person}{Greg Brockman},
  \bibinfo{person}{Alex Ray}, \bibinfo{person}{Raul Puri},
  \bibinfo{person}{Gretchen Krueger}, \bibinfo{person}{Michael Petrov},
  \bibinfo{person}{Heidy Khlaaf}, \bibinfo{person}{Girish Sastry},
  \bibinfo{person}{Pamela Mishkin}, \bibinfo{person}{Brooke Chan},
  \bibinfo{person}{Scott Gray}, \bibinfo{person}{Nick Ryder},
  \bibinfo{person}{Mikhail Pavlov}, \bibinfo{person}{Alethea Power},
  \bibinfo{person}{Lukasz Kaiser}, \bibinfo{person}{Mohammad Bavarian},
  \bibinfo{person}{Clemens Winter}, \bibinfo{person}{Philippe Tillet},
  \bibinfo{person}{Felipe~Petroski Such}, \bibinfo{person}{Dave Cummings},
  \bibinfo{person}{Matthias Plappert}, \bibinfo{person}{Fotios Chantzis},
  \bibinfo{person}{Elizabeth Barnes}, \bibinfo{person}{Ariel Herbert-Voss},
  \bibinfo{person}{William~Hebgen Guss}, \bibinfo{person}{Alex Nichol},
  \bibinfo{person}{Alex Paino}, \bibinfo{person}{Nikolas Tezak},
  \bibinfo{person}{Jie Tang}, \bibinfo{person}{Igor Babuschkin},
  \bibinfo{person}{Suchir Balaji}, \bibinfo{person}{Shantanu Jain},
  \bibinfo{person}{William Saunders}, \bibinfo{person}{Christopher Hesse},
  \bibinfo{person}{Andrew~N. Carr}, \bibinfo{person}{Jan Leike},
  \bibinfo{person}{Josh Achiam}, \bibinfo{person}{Vedant Misra},
  \bibinfo{person}{Evan Morikawa}, \bibinfo{person}{Alec Radford},
  \bibinfo{person}{Matthew Knight}, \bibinfo{person}{Miles Brundage},
  \bibinfo{person}{Mira Murati}, \bibinfo{person}{Katie Mayer},
  \bibinfo{person}{Peter Welinder}, \bibinfo{person}{Bob McGrew},
  \bibinfo{person}{Dario Amodei}, \bibinfo{person}{Sam McCandlish},
  \bibinfo{person}{Ilya Sutskever}, {and} \bibinfo{person}{Wojciech Zaremba}.}
  \bibinfo{year}{2021}\natexlab{}.
\newblock \showarticletitle{Evaluating Large Language Models Trained on Code}.
\newblock  (\bibinfo{year}{2021}).
\newblock
\showeprint[arxiv]{2107.03374}~[cs.LG]


\bibitem[Du et~al\mbox{.}(2023)]%
        {du-etal-2023-task}
\bibfield{author}{\bibinfo{person}{Chunhui Du}, \bibinfo{person}{Jidong Tian},
  \bibinfo{person}{Haoran Liao}, \bibinfo{person}{Jindou Chen},
  \bibinfo{person}{Hao He}, {and} \bibinfo{person}{Yaohui Jin}.}
  \bibinfo{year}{2023}\natexlab{}.
\newblock \showarticletitle{Task-Level Thinking Steps Help Large Language
  Models for Challenging Classification Task}. In
  \bibinfo{booktitle}{\emph{Proceedings of the 2023 Conference on Empirical
  Methods in Natural Language Processing}},
  \bibfield{editor}{\bibinfo{person}{Houda Bouamor}, \bibinfo{person}{Juan
  Pino}, {and} \bibinfo{person}{Kalika Bali}} (Eds.).
  \bibinfo{publisher}{Association for Computational Linguistics},
  \bibinfo{address}{Singapore}, \bibinfo{pages}{2454--2470}.
\newblock
\urldef\tempurl%
\url{https://doi.org/10.18653/v1/2023.emnlp-main.150}
\showDOI{\tempurl}


\bibitem[Entman(2007)]%
        {article}
\bibfield{author}{\bibinfo{person}{Robert Entman}.}
  \bibinfo{year}{2007}\natexlab{}.
\newblock \showarticletitle{Framing Bias: Media in the Distribution of Power}.
\newblock \bibinfo{journal}{\emph{Journal of Communication}}
  \bibinfo{volume}{57} (\bibinfo{date}{03} \bibinfo{year}{2007}),
  \bibinfo{pages}{163 -- 173}.
\newblock
\urldef\tempurl%
\url{https://doi.org/10.1111/j.1460-2466.2006.00336.x}
\showDOI{\tempurl}


\bibitem[Hendrycks et~al\mbox{.}(2020)]%
        {DBLP:journals/corr/abs-2009-03300}
\bibfield{author}{\bibinfo{person}{Dan Hendrycks}, \bibinfo{person}{Collin
  Burns}, \bibinfo{person}{Steven Basart}, \bibinfo{person}{Andy Zou},
  \bibinfo{person}{Mantas Mazeika}, \bibinfo{person}{Dawn Song}, {and}
  \bibinfo{person}{Jacob Steinhardt}.} \bibinfo{year}{2020}\natexlab{}.
\newblock \showarticletitle{Measuring Massive Multitask Language
  Understanding}.
\newblock \bibinfo{journal}{\emph{CoRR}}  \bibinfo{volume}{abs/2009.03300}
  (\bibinfo{year}{2020}).
\newblock
\showeprint[arXiv]{2009.03300}
\urldef\tempurl%
\url{https://arxiv.org/abs/2009.03300}
\showURL{%
\tempurl}


\bibitem[Hong et~al\mbox{.}(2023)]%
        {DBLP:conf/emnlp/HongCHJT23}
\bibfield{author}{\bibinfo{person}{Jiwoo Hong}, \bibinfo{person}{Yejin Cho},
  \bibinfo{person}{Jiyoung Han}, \bibinfo{person}{Jaemin Jung}, {and}
  \bibinfo{person}{James Thorne}.} \bibinfo{year}{2023}\natexlab{}.
\newblock \showarticletitle{Disentangling Structure and Style: Political Bias
  Detection in News by Inducing Document Hierarchy}. In
  \bibinfo{booktitle}{\emph{Findings of the Association for Computational
  Linguistics: {EMNLP} 2023, Singapore, December 6-10, 2023}},
  \bibfield{editor}{\bibinfo{person}{Houda Bouamor}, \bibinfo{person}{Juan
  Pino}, {and} \bibinfo{person}{Kalika Bali}} (Eds.).
  \bibinfo{publisher}{Association for Computational Linguistics},
  \bibinfo{pages}{5664--5686}.
\newblock
\urldef\tempurl%
\url{https://doi.org/10.18653/V1/2023.FINDINGS-EMNLP.377}
\showDOI{\tempurl}


\bibitem[Jiang et~al\mbox{.}(2021)]%
        {jiang2021convbertimprovingbertspanbased}
\bibfield{author}{\bibinfo{person}{Zihang Jiang}, \bibinfo{person}{Weihao Yu},
  \bibinfo{person}{Daquan Zhou}, \bibinfo{person}{Yunpeng Chen},
  \bibinfo{person}{Jiashi Feng}, {and} \bibinfo{person}{Shuicheng Yan}.}
  \bibinfo{year}{2021}\natexlab{}.
\newblock \bibinfo{title}{ConvBERT: Improving BERT with Span-based Dynamic
  Convolution}.
\newblock
\newblock
\showeprint[arxiv]{2008.02496}~[cs.CL]
\urldef\tempurl%
\url{https://arxiv.org/abs/2008.02496}
\showURL{%
\tempurl}


\bibitem[Lewis et~al\mbox{.}(2020)]%
        {lewis-etal-2020-bart}
\bibfield{author}{\bibinfo{person}{Mike Lewis}, \bibinfo{person}{Yinhan Liu},
  \bibinfo{person}{Naman Goyal}, \bibinfo{person}{Marjan Ghazvininejad},
  \bibinfo{person}{Abdelrahman Mohamed}, \bibinfo{person}{Omer Levy},
  \bibinfo{person}{Veselin Stoyanov}, {and} \bibinfo{person}{Luke
  Zettlemoyer}.} \bibinfo{year}{2020}\natexlab{}.
\newblock \showarticletitle{{BART}: Denoising Sequence-to-Sequence Pre-training
  for Natural Language Generation, Translation, and Comprehension}. In
  \bibinfo{booktitle}{\emph{Proceedings of the 58th Annual Meeting of the
  Association for Computational Linguistics}},
  \bibfield{editor}{\bibinfo{person}{Dan Jurafsky}, \bibinfo{person}{Joyce
  Chai}, \bibinfo{person}{Natalie Schluter}, {and} \bibinfo{person}{Joel
  Tetreault}} (Eds.). \bibinfo{publisher}{Association for Computational
  Linguistics}, \bibinfo{address}{Online}, \bibinfo{pages}{7871--7880}.
\newblock
\urldef\tempurl%
\url{https://doi.org/10.18653/v1/2020.acl-main.703}
\showDOI{\tempurl}


\bibitem[P{\'{e}}rez{-}Rosas et~al\mbox{.}(2018)]%
        {DBLP:conf/coling/Perez-RosasKLM18}
\bibfield{author}{\bibinfo{person}{Ver{\'{o}}nica P{\'{e}}rez{-}Rosas},
  \bibinfo{person}{Bennett Kleinberg}, \bibinfo{person}{Alexandra Lefevre},
  {and} \bibinfo{person}{Rada Mihalcea}.} \bibinfo{year}{2018}\natexlab{}.
\newblock \showarticletitle{Automatic Detection of Fake News}. In
  \bibinfo{booktitle}{\emph{Proceedings of the 27th International Conference on
  Computational Linguistics, {COLING} 2018, Santa Fe, New Mexico, USA, August
  20-26, 2018}}, \bibfield{editor}{\bibinfo{person}{Emily~M. Bender},
  \bibinfo{person}{Leon Derczynski}, {and} \bibinfo{person}{Pierre Isabelle}}
  (Eds.). \bibinfo{publisher}{Association for Computational Linguistics},
  \bibinfo{pages}{3391--3401}.
\newblock
\urldef\tempurl%
\url{https://aclanthology.org/C18-1287/}
\showURL{%
\tempurl}


\bibitem[Reid et~al\mbox{.}(2024)]%
        {DBLP:journals/corr/abs-2403-05530}
\bibfield{author}{\bibinfo{person}{Machel Reid}, \bibinfo{person}{Nikolay
  Savinov}, \bibinfo{person}{Denis Teplyashin}, \bibinfo{person}{Dmitry
  Lepikhin}, \bibinfo{person}{Timothy~P. Lillicrap},
  \bibinfo{person}{Jean{-}Baptiste Alayrac}, \bibinfo{person}{Radu Soricut},
  \bibinfo{person}{Angeliki Lazaridou}, \bibinfo{person}{Orhan Firat},
  \bibinfo{person}{Julian Schrittwieser}, \bibinfo{person}{Ioannis Antonoglou},
  \bibinfo{person}{Rohan Anil}, \bibinfo{person}{Sebastian Borgeaud},
  \bibinfo{person}{Andrew~M. Dai}, \bibinfo{person}{Katie Millican},
  \bibinfo{person}{Ethan Dyer}, \bibinfo{person}{Mia Glaese},
  \bibinfo{person}{Thibault Sottiaux}, \bibinfo{person}{Benjamin Lee},
  \bibinfo{person}{Fabio Viola}, \bibinfo{person}{Malcolm Reynolds},
  \bibinfo{person}{Yuanzhong Xu}, \bibinfo{person}{James Molloy},
  \bibinfo{person}{Jilin Chen}, \bibinfo{person}{Michael Isard},
  \bibinfo{person}{Paul Barham}, \bibinfo{person}{Tom Hennigan},
  \bibinfo{person}{Ross McIlroy}, \bibinfo{person}{Melvin Johnson},
  \bibinfo{person}{Johan Schalkwyk}, \bibinfo{person}{Eli Collins},
  \bibinfo{person}{Eliza Rutherford}, \bibinfo{person}{Erica Moreira},
  \bibinfo{person}{Kareem Ayoub}, \bibinfo{person}{Megha Goel},
  \bibinfo{person}{Clemens Meyer}, \bibinfo{person}{Gregory Thornton},
  \bibinfo{person}{Zhen Yang}, \bibinfo{person}{Henryk Michalewski},
  \bibinfo{person}{Zaheer Abbas}, \bibinfo{person}{Nathan Schucher},
  \bibinfo{person}{Ankesh Anand}, \bibinfo{person}{Richard Ives},
  \bibinfo{person}{James Keeling}, \bibinfo{person}{Karel Lenc},
  \bibinfo{person}{Salem Haykal}, \bibinfo{person}{Siamak Shakeri},
  \bibinfo{person}{Pranav Shyam}, \bibinfo{person}{Aakanksha Chowdhery},
  \bibinfo{person}{Roman Ring}, \bibinfo{person}{Stephen Spencer},
  \bibinfo{person}{Eren Sezener}, {and} \bibinfo{person}{et al.}}
  \bibinfo{year}{2024}\natexlab{}.
\newblock \showarticletitle{Gemini 1.5: Unlocking multimodal understanding
  across millions of tokens of context}.
\newblock \bibinfo{journal}{\emph{CoRR}}  \bibinfo{volume}{abs/2403.05530}
  (\bibinfo{year}{2024}).
\newblock
\urldef\tempurl%
\url{https://doi.org/10.48550/ARXIV.2403.05530}
\showDOI{\tempurl}
\showeprint[arXiv]{2403.05530}


\bibitem[Wang et~al\mbox{.}(2022)]%
        {DBLP:conf/emnlp/Wang0LTQYSHG22}
\bibfield{author}{\bibinfo{person}{Jianing Wang}, \bibinfo{person}{Chengyu
  Wang}, \bibinfo{person}{Fuli Luo}, \bibinfo{person}{Chuanqi Tan},
  \bibinfo{person}{Minghui Qiu}, \bibinfo{person}{Fei Yang},
  \bibinfo{person}{Qiuhui Shi}, \bibinfo{person}{Songfang Huang}, {and}
  \bibinfo{person}{Ming Gao}.} \bibinfo{year}{2022}\natexlab{}.
\newblock \showarticletitle{Towards Unified Prompt Tuning for Few-shot Text
  Classification}. In \bibinfo{booktitle}{\emph{Findings of the Association for
  Computational Linguistics: {EMNLP} 2022, Abu Dhabi, United Arab Emirates,
  December 7-11, 2022}}, \bibfield{editor}{\bibinfo{person}{Yoav Goldberg},
  \bibinfo{person}{Zornitsa Kozareva}, {and} \bibinfo{person}{Yue Zhang}}
  (Eds.). \bibinfo{publisher}{Association for Computational Linguistics},
  \bibinfo{pages}{524--536}.
\newblock
\urldef\tempurl%
\url{https://doi.org/10.18653/V1/2022.FINDINGS-EMNLP.37}
\showDOI{\tempurl}


\bibitem[Wei et~al\mbox{.}(2022)]%
        {DBLP:conf/nips/Wei0SBIXCLZ22}
\bibfield{author}{\bibinfo{person}{Jason Wei}, \bibinfo{person}{Xuezhi Wang},
  \bibinfo{person}{Dale Schuurmans}, \bibinfo{person}{Maarten Bosma},
  \bibinfo{person}{Brian Ichter}, \bibinfo{person}{Fei Xia},
  \bibinfo{person}{Ed~H. Chi}, \bibinfo{person}{Quoc~V. Le}, {and}
  \bibinfo{person}{Denny Zhou}.} \bibinfo{year}{2022}\natexlab{}.
\newblock \showarticletitle{Chain-of-Thought Prompting Elicits Reasoning in
  Large Language Models}. In \bibinfo{booktitle}{\emph{Advances in Neural
  Information Processing Systems 35: Annual Conference on Neural Information
  Processing Systems 2022, NeurIPS 2022, New Orleans, LA, USA, November 28 -
  December 9, 2022}}, \bibfield{editor}{\bibinfo{person}{Sanmi Koyejo},
  \bibinfo{person}{S.~Mohamed}, \bibinfo{person}{A.~Agarwal},
  \bibinfo{person}{Danielle Belgrave}, \bibinfo{person}{K.~Cho}, {and}
  \bibinfo{person}{A.~Oh}} (Eds.).
\newblock
\urldef\tempurl%
\url{http://papers.nips.cc/paper\_files/paper/2022/hash/9d5609613524ecf4f15af0f7b31abca4-Abstract-Conference.html}
\showURL{%
\tempurl}


\bibitem[Wen and Younes(2023)]%
        {Wen_2023}
\bibfield{author}{\bibinfo{person}{Zehao Wen} {and} \bibinfo{person}{Rabih
  Younes}.} \bibinfo{year}{2023}\natexlab{}.
\newblock \showarticletitle{ChatGPT v.s. media bias: A comparative study of
  GPT-3.5 and fine-tuned language models}.
\newblock \bibinfo{journal}{\emph{Applied and Computational Engineering}}
  \bibinfo{volume}{21}, \bibinfo{number}{1} (\bibinfo{date}{Oct.}
  \bibinfo{year}{2023}), \bibinfo{pages}{249–257}.
\newblock
\showISSN{2755-273X}
\urldef\tempurl%
\url{https://doi.org/10.54254/2755-2721/21/20231153}
\showDOI{\tempurl}


\bibitem[Wessel et~al\mbox{.}(2023)]%
        {Wessel2023}
\bibfield{author}{\bibinfo{person}{Martin Wessel}, \bibinfo{person}{Tomas
  Horych}, \bibinfo{person}{Terry Ruas}, \bibinfo{person}{Akiko Aizawa},
  \bibinfo{person}{Bela Gipp}, {and} \bibinfo{person}{Timo Spinde}.}
  \bibinfo{year}{2023}\natexlab{}.
\newblock \showarticletitle{Introducing MBIB - the first Media Bias
  Identification Benchmark Task and Dataset Collection}. In
  \bibinfo{booktitle}{\emph{Proceedings of 46th International ACM SIGIR
  Conference on Research and Development in Information Retrieval (SIGIR
  â€™23)}}. \bibinfo{publisher}{ACM}, \bibinfo{address}{New York, NY,
  USA}.
\newblock
\urldef\tempurl%
\url{https://doi.org/10.1145/3539618.3591882}
\showDOI{\tempurl}
\newblock
\shownote{ISBN 978-1-4503-9408-6/23/07}.


\end{thebibliography}










\end{document}